\documentclass[11pt,a4paper]{article}
\usepackage[hyperref]{acl2017}
\usepackage{times}
\usepackage{latexsym}
\usepackage{graphicx}
\usepackage{hyperref}
\usepackage{amsmath}
\usepackage{float}
\usepackage{amsfonts}
\usepackage{comment}
\usepackage{afterpage}
\usepackage{multirow}
\usepackage{array}
\usepackage{tabularx}
\usepackage{url}
\newcolumntype{L}[1]{>{\raggedright\let\newline\\\arraybackslash\hspace{0pt}}m{#1}}
\newcommand\numberthis{\addtocounter{equation}{1}\tag{\theequation}}
\usepackage{parallel,enumitem}
\usepackage{subcaption}
\usepackage[font=small,labelfont=bf]{caption}

\aclfinalcopy 

\title{Biomedical Event Trigger Identification Using Bidirectional Recurrent Neural Network Based Models}

\author{Patchigolla V S S Rahul, Sunil Kumar Sahu, Ashish Anand
\\ Department of Computer Science and Engineering \\ Indian Institute of Technology Guwahati, Assam, India 
\mbox{}\\
{\tt rahul.rahul.pvss@gmail.com}\\
{\tt \{sunil.sahu, anand.ashish\}@iitg.ernet.in}\\
}

\date{}

\begin{document}
\maketitle
\begin{abstract}
Biomedical events describe complex interactions between various biomedical entities. Event trigger is a word or a phrase which typically signifies the occurrence of an event. Event trigger identification is an important first step in all event extraction methods. However many of the current approaches either rely on complex hand-crafted features or consider features only within a window. In this paper we propose a method that takes the advantage of recurrent neural network (RNN) to extract higher level features present across the sentence. Thus hidden state representation of RNN along with word and entity type embedding as features avoid relying on the complex hand-crafted features generated using various NLP toolkits. Our experiments have shown to achieve state-of-art F1-score on Multi Level Event Extraction (MLEE) corpus. We have also performed category-wise analysis of the result and discussed the importance of various features in trigger identification task.
\end{abstract}
\section{Introduction}
Biomedical events play an important role in improving biomedical research in many ways. Some of its applications include pathway curation \citep{ohta2013overview} and development of domain specific semantic search engine \citep{ananiadou2015event}. So as to gain attraction among researchers many challenges such as BioNLP'09 \cite{kim2009overview}, BioNLP'11 \citep{kim2011overview}, BioNLP'13 \citep{nedellec2013overview} have been organized and many novel methods have also been proposed addressing these tasks.
\begin{center}
\begin{figure}[H]
\includegraphics[width=8cm]{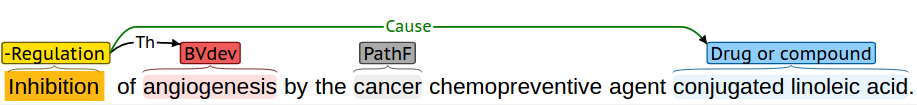}
\caption{Example of a complex biomedical event}
\end{figure}
\end{center}

An event can be defined as a combination of a trigger word and arbitrary number of arguments. Figure 1 shows two events with trigger words as ``\textit{Inhibition}'' and ``\textit{Angiogenesis}'' of trigger types ``\textit{Negative Regulation}'' and ``\textit{Blood Vessel Development}'' respectively. 
Pipelined based approaches for biomedical event extraction include event trigger identification followed by event argument identification. Analysis in multiple studies \citep{wang2016biomedical1,zhou2014event} reveal that more than 60\% of event extraction errors are caused due to incorrect trigger identification.

Existing event trigger identification models can be broadly categorized in two ways: \textit{rule based approaches} and \textit{machine learning based approaches}. Rule based approaches use various strategies including pattern matching and regular expression to define rules \citep{vlachos2009biomedical}. However, defining these rules are very difficult, time consuming and requires domain knowledge. The overall performance of the task depends on the quality of rules defined. These approaches often fail to generalize for new datasets when compared with machine learning based approaches. Machine learning based approaches treat the trigger identification problem as a word level classification problem, where many features from the data are extracted using various NLP toolkits \citep{pyysalo2012event,zhou2014event}  or learned automatically \citep{wang2016biomedical,wang2016biomedical1}. 

In this paper, we propose an approach using RNN to learn higher level features without the requirement of complex feature engineering. We thoroughly evaluate our proposed approach on MLEE corpus. We also have performed category-wise analysis and investigate the importance of different features in trigger identification task.

\section{Related Work}
Many approaches have been proposed to address the problem of event trigger identification. \citet{pyysalo2012event} proposed a model where various hand-crafted features are extracted from the processed data and fed into a Support Vector Machine (SVM) to perform final classification. \citet{zhou2014event} proposed a novel framework for trigger identification where embedding features of the word combined with hand-crafted features are fed to SVM for final classification using multiple kernel learning. \citet{wei2015hybrid} proposed a pipeline method on BioNLP'13 corpus based on Conditional Random Field (CRF) and Support vector machine (SVM) where the CRF is used to tag valid triggers and SVM is finally used to identify the trigger type. The above methods rely on various NLP toolkits to extract the hand-crafted features which leads to error propagation thus affecting the classifier's performance. These methods often need to tailor different features for different tasks, thus not making them generalizable. Most of the hand-crafted features are also traditionally sparse one-hot features vector which fail to capture the semantic information.

\citet{wang2016biomedical1} proposed a neural network model where dependency based word embeddings \citep{levy2014dependency} within a window around the word are fed into a feed forward neural network (FFNN) \cite{collobert11a} to perform final classification. \citet{wang2016biomedical} proposed another model based on convolutional neural network (CNN) where word and entity mention features of words within a window around the word are fed to a CNN to perform final classification. Although both of the methods have achieved good performance they fail to capture features outside the window.

\begin{center}
\begin{figure}
\includegraphics[scale=0.22]{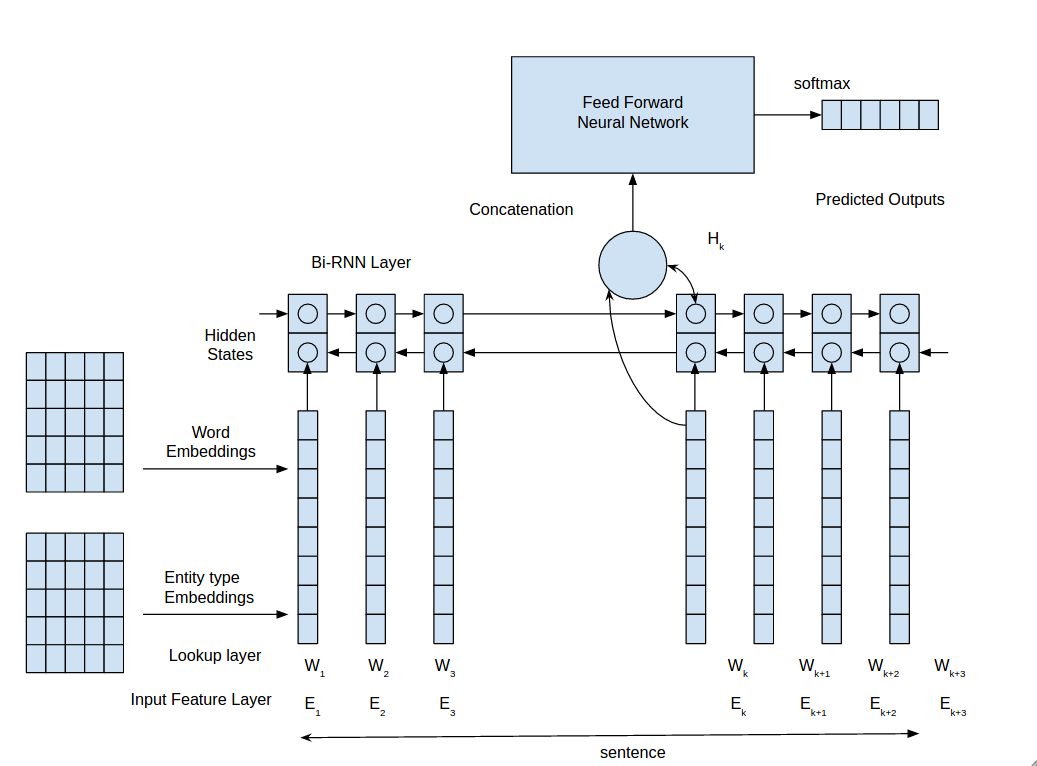}
\caption{Model Architecture}
\label{fig:model}
\end{figure}
\end{center}

\section{Model Architecture} 

We present our model based on bidirectional RNN as shown in Figure \ref{fig:model} for the trigger identification task. The proposed model detects trigger word as well as their type. Our model uses embedding features of words in the input layer and learns higher level representations in the subsequent layers and makes use of both the input layer and higher level features to perform the final classification. We now briefly explain about each component of our model.

\subsection{Input Feature Layer}
For every word in the sentence we extract two features, exact word $w$ $\in$ W and entity type $e$ $\in$ $E$. Here $W$ refers the dictionary of words and $E$ refers to dictionary of entities. Apart from all the entities, $E$ also contains a $None$ entity type which indicates absence of an entity. In some cases the entity might span through multiple words, in that case we assign every word spanned by that entity the same entity type.
\subsection{Embedding or Lookup Layer}
In this layer every input feature is mapped to a dense feature vector. Let us say that $E_w$ and $E_e$ be the embedding matrices of W and E respectively. The features obtained from these embedding matrices are concatenated and treated as the final word-level feature ($l$) of the model. 

The $E_w$ $\in \mathbb{R}^{n_w \times d_w}$ embedding matrix is initialized with pre-trained word embeddings and $E_e$ $\in \mathbb{R}^{n_e \times d_e}$ embedding matrix is initialized with random values. Here $n_w$, $n_e$ refer to length of the word dictionary and entity type dictionary respectively and $d_w$, $d_e$ refer to dimension of word and entity type embedding respectively.

\subsection{Bidirectional RNN Layer}
RNN is a powerful model for learning features from sequential data. We use both LSTM \citep{Hochreiter:1997:LSM:1246443.1246450} and GRU \citep{chung2014empirical} variants of RNN in our experiments as they handle the vanishing and exploding gradient problem \citep{pascanu2012understanding} in a better way. We use bidirectional version of RNN \citep{Graves13} where for every word forward RNN captures features from the past and the backward RNN captures features from future, inherently each word has information about whole sentence.

\subsection{Feed Forward Neural Network}
The hidden state of the bidirectional RNN layer acts as sentence-level feature ($g$), the word and entity type embeddings ($l$) act as a word-level features, are both concatenated \eqref{1.1} and passed through a series of hidden layers \eqref{1.2}, \eqref{1.3}  with dropout \citep{srivastava2014dropout} and an output layer. In the output layer, the number of neurons are equal to the number of trigger labels. Finally we use $Softmax$ function \eqref{1.4} to obtain probability score for each class.
\begin{align*}
f&=g^k \oplus l^k 				\numberthis \label{1.1}\\
h_0&=\tanh(W_0f+b_0)			\numberthis \label{1.2}	\\
h_i&=\tanh(W_ih_{i-1}+b_i)			\numberthis \label{1.3}	\\
p(y|x)&= Softmax(W_oh_i+b_o)	\numberthis \label{1.4}
\end{align*}
Here $k$ refers to the $k^{th}$ word of the sentence, $i$ refers to the $i^{th}$ hidden layer in the network and $\oplus$ refers to concatenation operation. $W_i$,$W_o$ and $b_i$,$b_o$ are parameters of the hidden and output layers of the network respectively.

\subsection{Training and Hyperparameters}
We use cross entropy loss function and the model is trained using stochastic gradient descent. The implementation\footnote{Implementation is available at \url{https://github.com/rahulpatchigolla/EventTriggerDetection}} of the model is done in python language using $Theano$ \citep{bergstra2010theano} library. We use pre-trained word embeddings obtained by
\citet{moen2013distributional} using \textit{word2vec} tool \citep{mikolov2013distributed}.

We use training and development set for hyperparameter selection. We use word embeddings of $200$ dimension, entity type embeddings of $50$ dimension, RNN hidden state dimension of $250$ and $2$ hidden layers with dimension $150$ and $100$. In both the hidden layers we use dropout of $0.2$.

\begin{figure*}[h!]
\begin{center}
\includegraphics[width=0.99\textwidth]{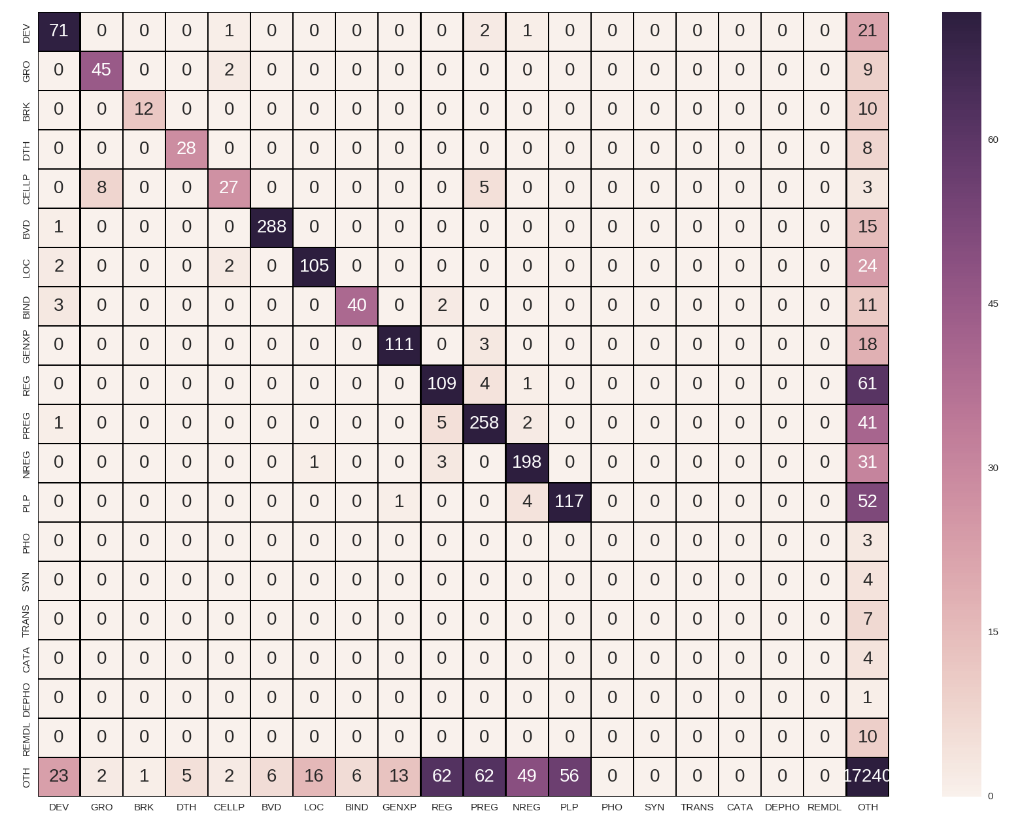}
\caption{Confusion matrix of trigger classes with abbreviations mentioned in Table \ref{table:5}}
\label{fig:conf}
\end{center}
\end{figure*}

\section{Experiments and discussion}
\label{sec:exp}
\subsection{Dataset Description}
\label{sec:dataset}
We use MLEE \citep{pyysalo2012event} corpus for performing our trigger identification experiments. Unlike other corpora on event extraction it covers events across various levels from molecular to organism level. The events in this corpus are broadly divided into 4 categories namely ``\textit{Anatomical}'', ``\textit{Molecular}'', ``\textit{General}'', ``\textit{Planned}'' which are further divided into 19 sub-categories as shown in Table \ref{table:5}. Here our task is to identify correct sub-category of an event. The entity types associated with the dataset are summarized in Table \ref{table:10}.
\begin{table}[H]
\resizebox{\columnwidth}{!}{%
\begin{tabular}{|l|l|l|l|}
\hline
\textbf{Category} & \textbf{Trigger label} & \textbf{Train count} & \textbf{Test count} \\
\hline
\multirow{7}{*}{Anatomical} & \multicolumn{1}{l|}{Cell Proliferation (CELLP)} & \multicolumn{1}{l|}{82} & \multicolumn{1}{l|}{43} \\\cline{2-4}
                                 & \multicolumn{1}{l|}{Development (DEV)} & \multicolumn{1}{l|}{202} & \multicolumn{1}{l|}{98} \\\cline{2-4}
                                 & \multicolumn{1}{l|}{Blood Vessel Development (BVD)} & \multicolumn{1}{l|}{540} & \multicolumn{1}{l|}{305}\\\cline{2-4}
                                 & \multicolumn{1}{l|}{Death (DTH)} & \multicolumn{1}{l|}{57} & \multicolumn{1}{l|}{36} \\\cline{2-4}
                                 & \multicolumn{1}{l|}{Breakdown (BRK)} & \multicolumn{1}{l|}{44} & \multicolumn{1}{l|}{23} \\\cline{2-4}
                                 & \multicolumn{1}{l|}{Remodeling (REMDL)} & \multicolumn{1}{l|}{22} & \multicolumn{1}{l|}{10} \\\cline{2-4}
                                 & \multicolumn{1}{l|}{Growth (GRO)} & \multicolumn{1}{l|}{107} & \multicolumn{1}{l|}{56}\\\hline
\multirow{6}{*}{Molecular} & \multicolumn{1}{l|}{Synthesis (SYN)} & \multicolumn{1}{l|}{13} & \multicolumn{1}{l|}{4}\\\cline{2-4}
                                 & \multicolumn{1}{l|}{Gene Expression (GENEXP)} & \multicolumn{1}{l|}{210} & \multicolumn{1}{l|}{132} \\\cline{2-4}
                                 & \multicolumn{1}{l|}{Transcription (TRANS)} & \multicolumn{1}{l|}{16} & \multicolumn{1}{l|}{7} \\\cline{2-4}
                                 & \multicolumn{1}{l|}{Catabolism (CATA)} & \multicolumn{1}{l|}{20} & \multicolumn{1}{l|}{4} \\\cline{2-4}
                                 & \multicolumn{1}{l|}{Phosphorylation (PHO)} & \multicolumn{1}{l|}{26} & \multicolumn{1}{l|}{3} \\\cline{2-4}
                                 & \multicolumn{1}{l|}{Dephosphorylation (DEPHO)} & \multicolumn{1}{l|}{2} & \multicolumn{1}{l|}{1}\\\hline
\multirow{5}{*}{General} & \multicolumn{1}{l|}{Localization (LOC)} & \multicolumn{1}{l|}{282} & \multicolumn{1}{l|}{133} \\\cline{2-4}
                                 & \multicolumn{1}{l|}{Binding (BIND)} & \multicolumn{1}{l|}{102} & \multicolumn{1}{l|}{56} \\\cline{2-4}
                                 & \multicolumn{1}{l|}{Regulation (REG)} & \multicolumn{1}{l|}{362} & \multicolumn{1}{l|}{178}\\\cline{2-4}
                                 & \multicolumn{1}{l|}{Positive Regulation (PREG)} & \multicolumn{1}{l|}{654} & \multicolumn{1}{l|}{312}\\\cline{2-4}
                                 & \multicolumn{1}{l|}{Negative Regulation (NREG)} & \multicolumn{1}{l|}{450} & \multicolumn{1}{l|}{233}\\\hline
Planned & Planned Process (PLP) &  407 & 175\\
\hline
\end{tabular}
}
\caption{Statistics of event triggers in MLEE corpus}
\label{table:5}
\end{table}
\begin{table}[H]
\resizebox{\columnwidth}{!}{%
\begin{tabular}{|l|l|l|l|}
\hline
\textbf{Category} & \textbf{Entity label} & \textbf{Train count} & \textbf{Test count} \\
\hline
\multirow{2}{*}{Molecule} & \multicolumn{1}{l|}{Drug or Compound} & \multicolumn{1}{l|}{637} & \multicolumn{1}{l|}{307}\\\cline{2-4}
                                 & \multicolumn{1}{l|}{Gene or Gene Product} & \multicolumn{1}{l|}{1961} & \multicolumn{1}{l|}{1001} \\\hline
\multirow{11}{*}{Anatomy} & \multicolumn{1}{l|}{Organism Subdivision} & \multicolumn{1}{l|}{27} & \multicolumn{1}{l|}{22} \\\cline{2-4}
                                 & \multicolumn{1}{l|}{Anatomical System} & \multicolumn{1}{l|}{10} & \multicolumn{1}{l|}{8} \\\cline{2-4}
                                 & \multicolumn{1}{l|}{Organ} & \multicolumn{1}{l|}{123} & \multicolumn{1}{l|}{53}\\\cline{2-4}
                                 & \multicolumn{1}{l|}{Multi-tissue Structure} & \multicolumn{1}{l|}{348} & \multicolumn{1}{l|}{166}\\\cline{2-4}
                                 & \multicolumn{1}{l|}{Tissue} & \multicolumn{1}{l|}{304} & \multicolumn{1}{l|}{122}\\\cline{2-4}
                                 & \multicolumn{1}{l|}{Cell} & \multicolumn{1}{l|}{866} & \multicolumn{1}{l|}{332}\\\cline{2-4}
                                 & \multicolumn{1}{l|}{Cellular Component} & \multicolumn{1}{l|}{105} & \multicolumn{1}{l|}{40}\\\cline{2-4}
                                 & \multicolumn{1}{l|}{Developing Anatomical Structure} & \multicolumn{1}{l|}{4} & \multicolumn{1}{l|}{2}\\\cline{2-4}
                                 & \multicolumn{1}{l|}{Organism Substance} & \multicolumn{1}{l|}{82} & \multicolumn{1}{l|}{60}\\\cline{2-4}
                                 & \multicolumn{1}{l|}{Immaterial Anatomical Entity} & \multicolumn{1}{l|}{11} & \multicolumn{1}{l|}{4}\\\cline{2-4}
                                 & \multicolumn{1}{l|}{Pathological Formation} & \multicolumn{1}{l|}{553} & \multicolumn{1}{l|}{357}\\\hline
Organism & Organism 
&  485 & 237\\
\hline
\end{tabular}
}
\caption{Statistics of entities in MLEE corpus}
\label{table:10}
\end{table}

\subsection{Experimental Design}
The data is provided in three parts as training, development and test sets. Hyperparameters are tuned using development set and then final model is trained on the combined set of training and development sets using the selected hyperparameters. The final results reported here are the best results over 5 runs.

We have used micro averaged F1-score as the evaluation metric and evaluated the performance of the model by ignoring the trigger classes with counts $\leq$ 10 in test set while training and considered them directly as false-negative while testing.

\subsection{Performance comparison with Baseline Models}
We compare our results with baseline models shown in Table \ref{table:3}. \citet{pyysalo2012event} defined a SVM based classifier with hand-crafted features. \citet{zhou2014event} also defined a SVM based classifier with word embeddings and hand-crafted features. \citet{wang2016biomedical} defined window based CNN classifier. Apart from the proposed models we also compare our results with two more baseline methods FFNN and CNN$^\psi$ which are our implementations. Here FFNN is a window based feed forward neural network where embedding features of words within the window are used to predict the trigger label \cite{collobert11a}. We chose window size as $3$ (one word from left and one word from right) after tuning it in validation set. CNN$^\psi$ is our implementation of window based CNN classifier proposed by \citet{wang2016biomedical} due to unavailability of their code in public domain. Our proposed model have shown slight improvement in F1-score when compared with baseline models. The proposed model's ability to capture the context of the whole sentence is likely to be one of the reasons of improvement in performance.

We perform one-side $t$-test over $5$ runs of F1-Scores to verify our proposed model's performance when compared with FFNN and CNN$^\Psi$. The $p$ value of the proposed model (GRU) when compared with FFNN and CNN$^\psi$ are $8.57\times 10^{-07}$ and $1.178\times 10^{-10}$ respectively. This indicates statistically superior performance of the proposed model.

\begin{table}[h]
\resizebox{\columnwidth}{!}{%
\begin{tabular}{|c|c|c|c|}
\hline
\textbf{Method} & \textbf{Precision} & \textbf{Recall} & \textbf{F1-Score}\\
\hline
SVM \citep{pyysalo2012event} & 81.44 & 69.48 & 75.67\\
\hline
SVM+$W_e$ \citep{zhou2014event} & 80.60 & 74.23 & 77.82\\
\hline
CNN \citep{wang2016biomedical} & 80.67 & 76.76 & 78.67\\
\hline
FFNN &77.53&75.55&76.53\\
\hline
CNN$^\psi$&80.75&69.36&74.62\\
\hline
Proposed (LSTM) & 78.58 & 78.84 & {\bf 78.71}\\
\hline
Proposed (GRU) & 79.78 & 78.45 & {\bf 79.11}\\
\hline
\end{tabular}
}
\caption{Comparison of performance of our model with baseline models}
\label{table:3}
\end{table}

\subsection{Category Wise Performance Analysis}
The category wise performance of the proposed model is shown in Table \ref{table:2}. It can be observed that model's performance in {\it anatomical} and {\it molecular} categories are better than {\it general} and {\it planned} categories. We can also infer from the confusion matrix shown in Figure \ref{fig:conf} that \textit{positive regulation}, \textit{negative regulation} and \textit{regulation} among {\it general} category and \textit{planned} category triggers are causing many false positives and false negatives thus degrading the model's performance.

\begin{table}[H]
\resizebox{\columnwidth}{!}{%
\begin{tabular}{|c|c|c|c|}
\hline
\textbf{Trigger Category} & \textbf{Precision} & \textbf{Recall} & \textbf{F1-Score}\\
\hline
Anatomical & 88.86&83.06&85.87\\
\hline
Molecular & 88.80 & 73.51& 80.43\\
\hline
General & 75.69&78.53&77.09\\
\hline
Planned & 67.63 & 67.24& 67.43\\
\hline
Overall & 79.78 & 78.45& 79.11\\
\hline
\end{tabular}
}
\caption{Category wise performance of the model}
\label{table:2}
\end{table}

\subsection{Further Analysis}

In this section we investigate the importance of various features and model variants as shown in Table \ref{table:4}. Here $E_w$ and $E_e$ refer to using word and entity type embedding as a feature in the model, $l$ and $g$ refer to using word-level and sentence-level feature respectively for the final prediction. For example, $E_w+E_e$ and $g$ means using both word and entity type embedding as the input feature for the model and $g$ means only using the global feature (hidden state of RNN) for final prediction.

\begin{table}[H]
\centering
\begin{tabular}{|c|c|c|}
\hline
\textbf{Index} & \textbf{Method} & \textbf{F1-Score}\\
\hline
1&$E_w$ and $g$&76.52\\
\hline
2&$E_w$ and $l+g$&77.59\\
\hline
3&$E_w+E_e$ and $g$&78.70\\
\hline
4&$E_w+E_e$ and $l+g$ & 79.11\\
\hline
\end{tabular}
\caption{Affect on F1-Score based on feature analysis and model variants}
\label{table:4}
\end{table}

Examples in Table \ref{table:ex_phrase} illustrate importance of features used in best performing models. In phrase 1 the word ``\textit{knockdown}'', is a part of an entity namely ``\textit{round about knockdown endothelial cells}'' of type ``\textit{Cell}'' and in phrase 2 it is trigger word of type ``\textit{Planned Process}'', methods 1 and 2 failed to differentiate both of them because of no knowledge about the entity type. In phrase 3 ``\textit{impaired}'' is a trigger word of type ``\textit{Negative Regulation}'' methods 1 and 3 failed to correctly identify but when reinforced with word-level feature the model succeeded in identification. So, we can say that $E_e$ feature and $l+g$ model variant help in improving the model's performance.

\begin{table}[H]
\centering
\begin{tabular}{|c|L{5.5cm}|}
\hline
\textbf{Index} & \textbf{Phrase}\\
\hline
1&silencing of directional migration in \textit{round about knockdown endothelial cells}\\
\hline
2&we show that PSMA inhibition \textit{knockdown} or deficiency decrease\\
\hline
3&display altered maternal hormone concentrations indicative of an \textit{impaired} trophoblast capacity\\
\hline
\end{tabular}
\caption{Example phrases for Further Analysis}
\label{table:ex_phrase}
\end{table}

\section{Conclusion and Future Work}
In this paper we have proposed a novel approach for trigger identification by learning higher level features using RNN. Our experiments have shown to achieve state-of-art results on MLEE corpus. In future we would like to perform complete event extraction using deep learning techniques.

\bibliography{acl2017}
\bibliographystyle{acl_natbib}

\appendix

\end{document}